# Vision Transformer for Robust Occluded Person Re-identification in Complex Surveillance Scenes

Bo Li[1,†], Duyuan Zheng[2,†], Xinyang Liu[2,†], Qingwen Li[1], Hong Li[1], Hongyan Cui[2,3,*],*Senior Member,IEEE*, Ge Gao[2], and Chen Liu[2]

[1] China Tower Corporation Limited, Beijing,100089, China
[2] China University of Petroleum-Beijing at Karamay, Karamay,834000, China
[3] Beijing University of Posts and Telecommunications, Beijing, 100876, China
[†] These authors contributed equally to this work and should be considered co-first authors.
*Corresponding author at:China University of Petroleum-Beijing at Karamay,Karamay,834000,China.
E-mail address:cuihy@bupt.edu.cn(H.Cui)

**Abstract.** Person re-identification (ReID) in surveillance is challenged by occlusion, viewpoint distortion, and poor image quality. Most existing methods rely on complex modules or perform well only on clear frontal images. We propose Sh-ViT (Shuffling Vision Transformer), a lightweight and robust model for occluded person ReID. Built on ViT-Base, Sh-ViT introduces three components: First, a Shuffle module in the final Transformer layer to break spatial correlations and enhance robustness to occlusion and blur; Second, scenario-adapted augmentation (geometric transforms, erasing, blur, and color adjustment) to simulate surveillance conditions; Third, DeiT-based knowledge distillation to improve learning with limited labels.To support real-world evaluation, we construct the MyTT dataset, containing over 10,000 pedestrians and 30,000+ images from base station inspections, with frequent equipment occlusion and camera variations. Experiments show that Sh-ViT achieves 83.2% Rank-1 and 80.1% mAP on MyTT, outperforming CNN and ViT baselines, and 94.6% Rank-1 and 87.5% mAP on Market1501, surpassing state-of-the-art methods.In summary, Sh-ViT improves robustness to occlusion and blur without external modules, offering a practical solution for surveillance-based personnel monitoring.

## 1 Introduction

Person re-identification (ReID) [1] is a fundamental problem in computer vision, aiming to identify the same pedestrian across different camera views. It has broad applications in public safety, intelligent surveillance, and operation & maintenance (O&M) management.

In base station O&M scenarios, the goal of ReID is to distinguish authorized from unauthorized personnel. As shown in Fig. 1, the base station scenario has unique imaging characteristics and challenges: First, surveillance cameras are typically installed at elevated or oblique angles, leading to noticeable pose tilt and geometric distortion. Second, blur from poor lighting or hardware and frequent equipment-person occlusion further reduce fine-grained details. Third, in complex working environments, frequent interactions between personnel and equipment often result in partial or large-scale occlusion. Tilt, blur, and occlusion are the main bottlenecks of ReID in O&M applications, and form the focus of this study.

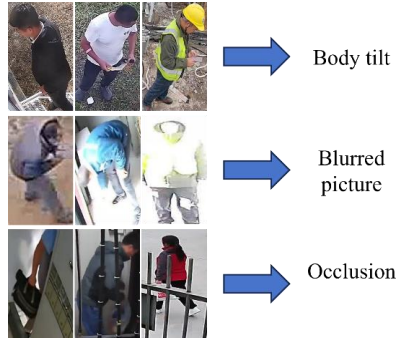

**Fig. 1.** Person re-identification problem, showing the limitations of convolutional kernel window sliding for feature extraction in complex scenes (including person tilt, illumination change, background interference), and the impact of strong data dependence and limited generalization ability of CNN on feature extraction robustness/accuracy and character recognition accuracy.

To address these issues, we propose Shuffled Vision Transformer (Sh-ViT), a lightweight yet robust framework for occluded person ReID. Sh-ViT incorporates three key innovations: First, a Shuffle module that randomly rearranges feature tokens to enhance robustness against occlusion and blurring; Second, scenario-adapted data augmentation that simulates top-down views, equipment occlusions, lighting variations, and blur degradations; Third, knowledge distillation from DeiT, improving stability and data efficiency under limited annotations.

In addition, we construct a new dataset, MyTT, containing over 10,000 pedestrians and 30,000+ images collected from real base station scenarios, which highlights challenges not well captured by public datasets.Extensive experiments demonstrate that Sh-ViT achieves 83.2% Rank-1 and 80.1% mAP on MyTT,94.6% Rank-1 and 87.5% mAP on Market1501, surpassing recent state-of-the-art methods. These results confirm that Sh-ViT offers a practical solution for person ReID in complex O&M surveillance scenarios.

The remainder of this paper is organized as follows: Section II reviews deep learning-based feature extraction methods. Section III introduces the proposed Sh-ViT model. Section IV presents experimental results and comparisons with existing ViT models. Section V summarizes the work presented in this paper.

## 2    Related Work

### 2.1    Occluded Person Re-Identification

Occluded person ReID is important in real surveillance. Traditional methods relied on part-based pooling or parsing (PCB-RPP [2], HOReID [3]). With Transformers, attention-based recovery has been studied, such as feature completion (FCFormer [4]) or dynamic patch selection (DPEFormer [5]). However, these methods depend on synthetic occlusion or complex modules. In contrast, our Shuffle mechanism directly enhances global robustness without explicit occlusion detection.

### 2.2    Vision Transformers in ReID

Transformers have been adapted for ReID, e.g., TransReID [6] with side information, self-supervised pre-training [7], and masked image modeling (PersonViT [8]). These works mainly address holistic recognition but lack designs for severe occlusion. Our Sh-ViT extends ViT with a shuffle module, improving tolerance to missing body regions.

### 2.3    Data Augmentation for Robust ReID

Conventional augmentations (e.g., random erasing [9]) and occlusion simulation [10] help robustness, but they overlook surveillance-specific challenges such as equipment occlusion and perspective distortion. We design scenario-adapted augmentation combining geometric transforms, realistic occlusion, blur, and lighting adjustments to better match real scenes.

### 2.4    Knowledge Distillation in ReID

Distillation improves ReID efficiency and performance. DeiT [11] introduced distillation tokens, while later works [12] adapted it for ReID tasks. We adopt DeiT distillation to enhance data efficiency, which is critical when labeled data are limited.

## 3    The Sh-ViT Method

This section introduces the proposed Shuffled Vision Transformer (Sh-ViT) for person Re-ID in complex surveillance scenarios. Sh-ViT is designed to address three key challenges—viewpoint distortion, image degradation, and occlusion—which frequently occur in base station inspection environments. As illustrated in Fig. 2, the framework is built upon a ViT-Base backbone and integrates three major components: a Shuffle module, scenario-adapted data augmentation, and knowledge distillation.

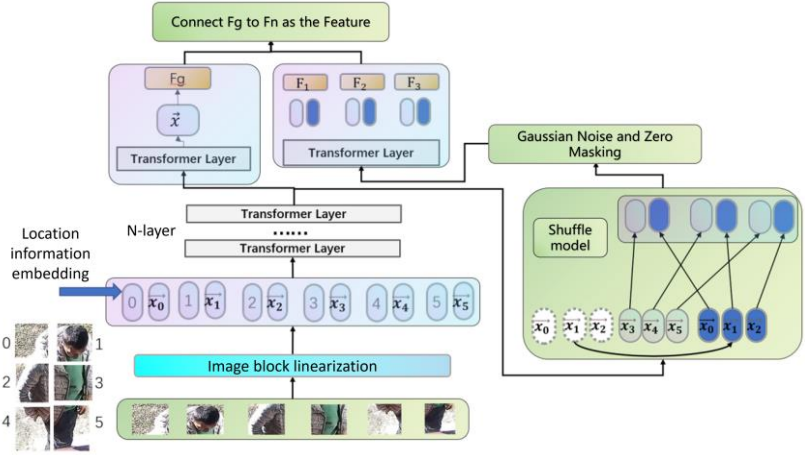

**Fig. 2.** Structure of Re-ID model. Structure of Shuffled Vision Transformer (Sh-ViT) for person re-identification, illustrating image patch linearization, location information embedding, multi-layer Transformer operations, shuffle module with Gaussian noise and zero masking, and feature concatenation of $F_g$ and $F_n$.

### 3.1    Overall Framework

The overall pipeline of Sh-ViT follows the standard Transformer-based architecture for visual representation learning. Input images are divided into non-overlapping patches and mapped to high-dimensional embeddings, which are processed by a multi-layer Transformer encoder to capture long-range dependencies. To enhance robustness under practical conditions, Sh-ViT incorporates three innovations: First, a Shuffle module embedded in the last Transformer layer to improve resistance to occlusion and blur; Second, a scenario-adapted data augmentation strategy tailored to surveillance environments; Third, DeiT-based knowledge distillation to improve data efficiency. The final representation is normalized and fed into a fully connected classifier for identity prediction.

### 3.2    Shuffle Module

The Shuffle module distinguishes itself from existing patch-shuffling or occlusion-handling techniques in both conceptual design and practical implementation. Conceptually, unlike methods that rely on explicit occlusion reconstruction (e.g., FCFormer [4], which uses a complex decoder to recover missing occluded regions) or localized patch enhancement (e.g., PHA [19], which only augments high-frequency patches), the Shuffle module does not require prior occlusion detection or synthetic mask generation. Instead, it forces the model to learn spatially invariant global features by randomly permuting patch tokens, thus avoiding over-reliance on specific occlusion patterns (e.g., the fixed synthetic occlusions in [10]). Practically, in contrast to general patch-shuffling techniques applied in input preprocessing (e.g., the random erasing-based shuffling in [9]), our Shuffle module is embedded in the last layer of the Transformer encoder and integrates Gaussian noise injection and zero masking. This design ensures that the model retains high-level semantic information while enhancing robustness—existing input-level shuffling methods often disrupt spatial logic and degrade the discriminability of local features, leading to performance loss in identity recognition tasks.

The Shuffle module is the core contribution of Sh-ViT and aims to mitigate the adverse effects of partial occlusion and image blur.

●Token shuffling: Patch tokens are randomly grouped and permuted, breaking local spatial correlations and forcing the model to capture global dependencies across visible fragments.

●Reconstruction through attention: The shuffled tokens are reassembled via the self-attention mechanism, enabling the model to reconstruct robust identity features even when large portions of the body are missing.

●Robustness enhancement: Gaussian noise and zero masking are incorporated into the shuffled tokens, improving tolerance to image degradation and missing information.

Unlike input-level patch shuffling [9] or occlusion simulation [10], our shuffle module is embedded in the final Transformer layer. This design avoids semantic distortion and compels the model to learn spatially invariant global features.

This lightweight design introduces negligible computational overhead while substantially improving feature robustness under occlusion-heavy and blur-prone conditions.

### 3.3 Scenario-Adapted Data Augmentation

We design scenario-adapted augmentation (affine/perspective transforms, random erasing, Gaussian blur, and color adjustment) to simulate surveillance distortions.

●Affine and perspective transformations to simulate geometric distortion caused by elevated or tilted cameras;

●Random erasing to mimic occlusions from equipment or other personnel;

●Gaussian blur to handle motion blur and out-of-focus images;

●Color adjustment to address variations in illumination, saturation, and contrast.

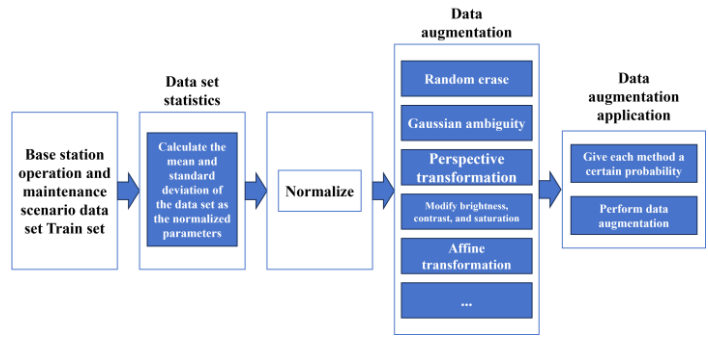

**Fig. 3.** Pipeline of Data Augmentation. Pipeline of data augmentation for base station operation and maintenance scenario dataset, showing steps of dataset statistics (calculating mean and standard deviation as normalization parameters), normalization, diverse augmentation methods (random erase, Gaussian ambiguity, perspective transformation, brightness/contrast/saturation modification, affine transformation, etc.), and probabilistic application of data augmentation.

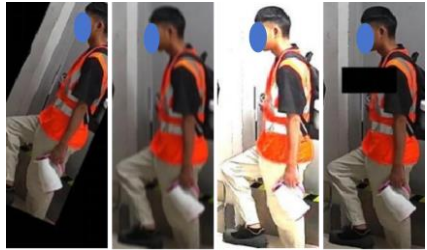

**Fig. 4.** Examples of different data augmentation methods (Transform, Random Erase, Gaussian Blur, and Color transform) applied to personnel images in base station operation and maintenance scenarios.

### 3.4    Self-Built MyTT Dataset

For base station inspection scenarios, we construct the MyTT dataset from real multi-camera surveillance videos, formatted to match Re-ID task standards (e.g., Market1501). Key details:

●Scale: 10,000+ pedestrians, 30,000+ images; split into training/test/validation sets at 6:3:1.

●Scenario Features: Top-down shooting angles, frequent equipment-person occlusion, unstable lighting, cross-camera color/clarity differences, and partial image blurriness—addressing gaps in public datasets (e.g., Market1501 [13]) that lack complex inspection-scene characteristics.

## 4      Experiments

### 4.1    Experimental Setups

To verify the effectiveness of Sh-ViT, experiments are conducted on the self-built occluded person Re-ID dataset MyTT and the mainstream benchmark Market1501 [14].

**MyTT:** Constructed for security equipment room Re-ID, collected from multi-camera surveillance. It contains over 10,000 pedestrians and 30,000+ images, split into training (6,654 identities, 34,536 images), test (4,437 identities, 17,359 images), and validation (4,206 identities, 5,786 images) sets, formatted to match Market1501.

**Market1501:** A common holistic Re-ID benchmark captured by 6 cameras, with 751 training identities (12,936 images) and 750 test identities (19,732 query + 3,368 gallery images).

### 4.2    Comparative Experiments

We propose Sh-ViT by integrating a Shuffle module into the ViT-based Transformer, aiming to improve ViT's performance in occluded scenarios. Table 1 compares Sh-ViT with state-of-the-art (SOTA) methods on MyTT. Sh-ViT achieves 83.2% Rank-1 and 80.1% mAP, outperforming conventional CNNs (e.g., ResNet50: 82.1% Rank-1, 57.6% mAP; OSNet: 72.7% Rank-1, 54.2% mAP) and the ViT-based baseline

TransReID (83.0% Rank-1, 79.6% mAP). This confirms Sh-ViT's superiority in handling occlusions and low-quality images in security inspection scenarios.

**Table 1.** Superiority of Sh-ViT on MyTT Dataset.

| Method | Publication | Rank1(%) | Rank5 (%) | Rank10 (%) | mAP(%) |
|---|---|---|---|---|---|
| OSNet [14] | ICCV2019 | 72.7 | 85.0 | 87.9 | 54.2 |
| ABDNet [15] | ICCV2019 | 76.2 | 86.0 | 89.8 | 57.1 |
| HOReID [3] | CVPR2020 | 69.6 | 85.7 | 88.6 | 53.9 |
| TransReID [6] | ICCV2021 | 83.0 | 84.2 | 87.8 | 79.6 |
| ResNet50 [16] | CVPR2016 | 82.1 | 86.5 | 90.3 | 57.6 |
| ResNeSt50 [17] | CVPR2022 | 78.6 | 88.1 | 89.5 | 58.7 |
| MSINet [18] | CVPR2023 | 72.5 | 89.7 | 92.7 | 59.7 |
| PHA [19] | CVPR2023 | 76.9 | 89.6 | 91.5 | 58.8 |
| UniHCP [20] | CVPR2023 | 76.5 | 92.5 | 94.6 | 57.6 |
| DeepChange [21] | ICCV2023 | 70.8 | 82.2 | 86.0 | 46.6 |
| **Sh-ViT (Ours)** | **This paper** | **83.2** | 92.7 | 94.7 | **80.1** |

### 4.3 Cross-Domain Generalization Validation

To evaluate Sh-ViT's generalization, we test it on Market1501 (Table 2). Sh-ViT achieves 94.6% Rank-1, 98.2% Rank-5, 99.1% Rank-10, and 87.5% mAP, outperforming recent SOTA methods such as DCReID (94.2% Rank-1, 85.5% mAP) and ACFL (94.3% Rank-1, 85.3% mAP). This gain is attributed to the Shuffle module, which enhances the model's ability to mitigate occlusion interference and capture robust global features, enabling adaptation to different surveillance scenarios.

**Table 2.** Performance of Sh-ViT on Market1501.

| Method | Publication | Rank1(%) | Rank5(%) | Rank10(%) | mAP(%) |
|---|---|---|---|---|---|
| D-MMD [22] | ECCV2020 | 70.6 | 87.0 | 91.5 | 48.8 |
| MMCL [23] | CVPR2020 | 84.4 | 92.8 | 95.0 | 60.4 |
| MLC [24] | PR2022 | 85.6 | 93.9 | 96.0 | 65.9 |
| MDJL [25] | PR2023 | 80.3 | 87.4 | 89.9 | 59.8 |
| AdaMG [26] | TCSVT2023 | 93.9 | 97.9 | 98.9 | 84.6 |
| ACFL [27] | PR2024 | 94.3 | 98.0 | 98.8 | 85.3 |
| FCM [28] | AAAI2024 | 93.2 | 96.7 | 97.6 | 83.5 |
| DCReID [29] | CVDL2024 | 94.2 | 97.7 | 98.5 | 85.5 |
| **Sh-ViT (Ours)** | **This paper** | **94.6** | **98.2** | **99.1** | **87.5** |

### 4.4 Performance on Occlusion-Specific Benchmark (DukeMTMC-reID)

We further evaluate Sh-ViT on the occlusion-specific benchmark DukeMTMC-reID. As shown in Table 3, Sh-ViT achieves 89.6% Rank-1 and 80.3% mAP, surpassing recent methods such as ACFL [27] (85.2% Rank-1/76.8% mAP) and FCM [28] (84.7%

Rank-1/75.9% mAP). These results confirm the strong generalization ability of Sh-ViT under real-world occlusion conditions.

**Table 3.** Performance comparison with state-of-the-art models on DukeMTMC-reID.

| Method | Publication | Rank1(%) | Rank5 (%) | Rank10 (%) | mAP(%) |
|---|---|---|---|---|---|
| D-MMD [22] | ECCV2020 | 63.5 | 78.8 | 83.9 | 46.0 |
| MMCL [23] | CVPR2020 | 72.4 | 82.9 | 85.0 | 51.4 |
| MLC [24] | PR2022 | 74.1 | 83.8 | 86.3 | 55.0 |
| MDJL [25] | PR2023 | 78.6 | 86.6 | 88.7 | 62.8 |
| ACFL [27] | PR2024 | 85.5 | 92.4 | 94.4 | 74.0 |
| FCM [28] | AAAI2024 | 83.8 | 91.0 | 93.2 | 71.9 |
| **Sh-ViT(Ours)** | **This papaer** | **89.6** | 95.5 | 96.8 | **80.3** |

### 4.5    Algorithm Analysis

We analyze Sh-ViT's key components (data augmentation, backbone, optimizer) based on sufficient dataset images (about 30,000 images).

#### 4.5.1    Effect of Data Augmentation

Table 4 compares ViT performance with and without the proposed scenario-adapted augmentation (affine transformation, random erasing, Gaussian blur, color adjustment). Augmentation improves Rank-1 by +8.2% with limited data, but has little effect with sufficient data.

**Table 4.** Results of Data Augmentation Experiment.

| Method | Rank1(%) | Rank5(%) | Rank10(%) | mAP(%) |
|---|---|---|---|---|
| ViT (Insufficient data) | 60.9 | 86.1 | 90.1 | 68.7 |
| ViT + Data Augmentation (Insufficient data) | 69.1 | 86.4 | 91.2 | 68.9 |
| ViT (Sufficient data) | 83.0 | 92.5 | 94.6 | 79.6 |
| ViT + Data Augmentation (Sufficient data) | 82.8 | 92.4 | 94.2 | 79.1 |

#### 4.5.2    Effect of Backbone and Shuffle module

Table 5 compares ViT, DeiT, and Sh-ViT on MyTT. ViT (83.0% Rank-1, 79.6% mAP) outperforms DeiT (82.7% Rank-1, 78.7% mAP), so ViT is selected as the backbone. Adding the Shuffle module (Sh-ViT) further improves Rank-1 to 83.2% and stabilizes mAP, verifying the module's ability to enhance occlusion robustness.

**Table 5.** Results of Backbone and Shuffle module Experiment.

| Method | Rank1(%) | Rank5(%) | Rank10(%) | mAP(%) |
|---|---|---|---|---|
| DeiT | 82.7 | 92.3 | 94.3 | 78.7 |
| ViT | 83.0 | 92.5 | 94.6 | 79.6 |
| **Sh-ViT (Ours)** | **83.2** | **92.7** | **94.7** | **80.1** |

### 4.5.3 Effect of Optimizer

Table 6 compares SGD, Adan, and Adam on ViT. SGD achieves the best performance, while Adan and Adam underperform. SGD achieves the best performance and is adopted.

**Table 6.** Results of Optimizer Experiment.

| Method | Rank1(%) | Rank5(%) | Rank10(%) | mAP(%) |
|---|---|---|---|---|
| ViT + SGD | 82.8 | 92.5 | 94.4 | 79.4 |
| ViT + Adan | 77.3 | 88.4 | 91.0 | 68.9 |
| ViT + Adam | 72.8 | 83.7 | 86.7 | 65.6 |

## 4.6 Summary of Experiments

Our experiments across both public and self-built datasets confirm the robustness and effectiveness of Sh-ViT for occluded person re-identification under complex surveillance conditions.

**Performance on Public Benchmarks.** On Market1501, Sh-ViT achieves 94.6% Rank-1 and 87.5% mAP, outperforming recent state-of-the-art approaches such as DCReID and ACFL. The consistent gains across Rank-5 and Rank-10 show that the Shuffle module enhances global feature robustness beyond the training domain. These improvements are particularly notable considering that Market1501 is less occlusion-heavy than our target scenarios, suggesting Sh-ViT generalizes well even in less constrained environments.

**Evaluation on the MyTT Dataset.** The MyTT dataset, designed to replicate real-world base station inspections, poses unique challenges: severe occlusion from equipment, top-down camera angles, unstable lighting, and cross-camera style shifts. On MyTT, Sh-ViT reaches 83.2% Rank-1 and 80.1% mAP, surpassing CNN-based baselines such as ResNet5 and OSNet. The results indicate that ViT's global attention combined with patch-level shuffling significantly improves resistance to partial occlusion and low image quality.

**Effectiveness of Model Components.** Ablation studies show that patch shuffling contributes directly to performance gains by breaking local correlations and forcing the Transformer to learn spatially invariant features. Although the performance increase from ViT to Sh-ViT on MyTT appears modest, the Shuffle module notably stabilizes mAP and improves robustness under severe occlusion (as observed in qualitative retrieval results). Scenario-specific augmentation—particularly random erasing and Gaussian blur—proves beneficial when labeled data are scarce, improving Rank-1 by 8.2% under limited-data conditions. Furthermore, the optimizer comparison demonstrates that SGD is more stable and effective than Adan or Adam in this setting, aligning with prior findings for ReID tasks.

**Cross-Domain Generalization.** The competitive performance on both Market1501 and MyTT validates Sh-ViT's ability to adapt across domains without relying on external parsing networks or pose estimators. Typical failure cases occur under heavy equipment occlusion (over 60% body covered) or extreme illumination, where performance drops significantly. These cases suggest directions for future refinement.

**Limitations and Future Work.** Despite its strengths, Sh-ViT still shows only incremental gains over TransReID on some metrics, indicating room for further refinement. Main failures are heavy occlusion and extreme lighting; future work may adopt adaptive token weighting and stronger illumination augmentation. The MyTT dataset, while diverse, focuses on base station environments; broader evaluation on additional occluded benchmarks could provide a more comprehensive assessment. Future work may explore adaptive shuffling strategies or integrating temporal cues from multi-frame sequences to enhance tracking stability in video-based ReID.

## 5    Conclusion

In this paper, we proposed Sh-ViT, a Vision Transformer framework tailored for occluded person re-identification in surveillance. Sh-ViT combines a shuffle module, scenario-adapted augmentation, and DeiT-based distillation, delivering strong robustness with lightweight design.

Experiments confirm its effectiveness: Sh-ViT reaches 83.2% Rank-1 / 80.1% mAP on MyTT, 94.6% / 87.5% on Market1501, and 89.6% / 80.3% on Occluded-Duke, showing consistent superiority and strong cross-dataset generalization. Nevertheless, gains over the TransReID baseline are incremental on some metrics, indicating room for further refinement.

Future work will explore adaptive shuffle strategies that dynamically adjust permutation intensity according to occlusion patterns, and temporal modeling for video-based ReID. Preliminary tests on MARS already suggest temporal extension can further improve performance, making Sh-ViT promising for real-time surveillance applications.

**Acknowledgments.** This work was supported by National Natural Science Foundation of China (6217104), China University of Petroleum (Beijing) Karamay Campus introduction of talents and launch of scientific research projects(XQZX20240010), and China Tower Corporation Limited IT System 2023 Package Software Project - AI Algorithm and Services(23M01ZBZB011000017).

# References

1. Zheng, Liang, Yi Yang, and Alexander G. Hauptmann. "Person re-identification: Past, present and future." arXiv preprint arXiv:1610.02984 (2016)
2. Sun, Y., Zheng, L., Yang, Y., Tian, Q., Wang, S.: Beyond Part Models: Person Retrieval with Refined Part Pooling. In: Ferrari, V., Hebert, M., Sminchisescu, C., Weiss, Y. (eds.) ECCV 2018, LNCS, vol. 11205, pp. 480-496. Springer, Cham (2018)
3. H. Huang, D. Chen, X. Li, S. Wang, Z. Lei: HOReID: Deep High-Order Mapping for Occluded Person Re-Identification. In: Proc. IEEE/CVF Conference on Computer Vision and Pattern Recognition (CVPR), pp. 3218–3227. IEEE Press, New York (2020)
4. Wang, T., Liu, H., Song, P., Sun, T., Jin, Y.: Feature Completion Transformer for Occluded Person Re-Identification. IEEE Transactions on Multimedia 26, 8529–8542 (2024)
5. Zhang, X., et al.: Dynamic Patch-aware Enrichment Transformer for Occluded Person Re-Identification. arXiv preprint arXiv:2402.10435 (2024)
6. He, S., Luo, H., Wang, P., Wang, F., Li, H., Jiang, W.: TransReID: Transformer-based Object Re-Identification. In: 2021 IEEE/CVF International Conference on Computer Vision (ICCV), pp. 14993–15002 (2021)
7. Luo, H., et al.: Self-Supervised Pre-Training for Transformer-Based Person Re-Identification. arXiv preprint arXiv:2111.12084 (2021)
8. Hu, B., et al.: PersonViT: Large-scale Self-supervised Vision Transformer for Person Re-Identification. Machine Vision and Applications 36(1), 32 (2024)
9. Zhong, Z., Zheng, L., Kang, G., Li, S., Yang, Y.: Random Erasing Data Augmentation. In: Proceedings of the AAAI Conference on Artificial Intelligence, vol. 34, no. 07, pp. 13001–13008 (2020)
10. Wang, Y., et al.: Occlusion-aware R-CNN: Detecting Pedestrians in the Wild with Synthetic Occlusion. In: 2018 IEEE/CVF Conference on Computer Vision and Pattern Recognition Workshops (CVPRW), pp. 639–647 (2018)
11. Touvron, H., et al.: Training data-efficient image transformers & distillation through attention. In: Proceedings of the 37th International Conference on Machine Learning. PMLR, vol. 119, pp. 10347–10357 (2020)
12. Yang, J., et al.: Knowledge distillation via adaptive instance normalization. arXiv preprint arXiv:2003.04289 (2020)
13. Zheng, L., Shen, L., Tian, L., Wang, S., Wang, J., Tian, Q.: Scalable Person Re-identification: A Benchmark. In: 2015 IEEE International Conference on Computer Vision (ICCV), pp. 1116–1124 (2015)
14. Zhou, K., Yang, Y., Cavallaro, A., Xiang, T.: Omni-Scale Feature Learning for Person Re-Identification. In: 2019 IEEE/CVF International Conference on Computer Vision (ICCV), pp. 3701–3711 (2019)
15. Chen, T., et al.: ABD-Net: Attentive but Diverse Person Re-Identification. In: 2019 IEEE/CVF International Conference on Computer Vision (ICCV), pp. 8350–8360 (2019)
16. He, K., Zhang, X., Ren, S., Sun, J.: Deep Residual Learning for Image Recognition. In: 2016 IEEE Conference on Computer Vision and Pattern Recognition (CVPR), pp. 770–778 (2016)
17. Zhang, H., et al.: ResNeSt: Split-Attention Networks. In: 2022 IEEE/CVF Conference on Computer Vision and Pattern Recognition Workshops (CVPRW), pp. 2735–2745 (2022)
18. Gu, J., et al.: MSINet: Twins Contrastive Search of Multi-Scale Interaction for Object ReID. In: 2023 IEEE/CVF Conference on Computer Vision and Pattern Recognition (CVPR), pp. 19243–19253 (2023)

19. Zhang, G., Zhang, Y., Zhang, T., Li, B., Pu, S.: PHA: Patch-Wise High-Frequency Augmentation for Transformer-Based Person Re-Identification. In: 2023 IEEE/CVF Conference on Computer Vision and Pattern Recognition (CVPR), pp. 14133–14142 (2023)

20. Ci, Y., et al.: UniHCP: A Unified Model for Human-Centric Perceptions. In: 2023 IEEE/CVF Conference on Computer Vision and Pattern Recognition (CVPR), pp. 17840–17852 (2023)

21. Xu, P., Zhu, X.: DeepChange: A Long-Term Person Re-Identification Benchmark with Clothes Change. In: 2023 IEEE/CVF International Conference on Computer Vision (ICCV), pp. 11162–11171 (2023)

22. Mekhazni, D., Bhuiyan, A., Ekladious, G., Granger, E.: Unsupervised Domain Adaptation in the Dissimilarity Space for Person Re-identification. In: Vedaldi, A., Bischof, H., Brox, T., Frahm, J.-M. (eds.) ECCV 2020. LNCS, vol. 12370, pp. 159–174. Springer, Cham (2020)

23. Wang, D., Zhang, S.: Unsupervised Person Re-Identification via Multi-Label Classification. In: 2020 IEEE/CVF Conference on Computer Vision and Pattern Recognition (CVPR), pp. 10978–10987 (2020)

24. Li, Q., et al.: Unsupervised Person Re-Identification with Multi-Label Learning Guided Self-Paced Clustering. arXiv preprint arXiv:2103.04580 (2021)

25. Chen, F., Wang, N., Tang, J., Yan, P., Yu, J.: Unsupervised Person Re-Identification via Multi-Domain Joint Learning. Pattern Recognition 138, 109369 (2023)

26. Peng, J., Jiang, G., Wang, H.: Adaptive Memorization with Group Labels for Unsupervised Person Re-Identification. IEEE Transactions on Circuits and Systems for Video Technology 33(10), 5802–5813 (2023)

27. Ji, H., Wang, L., Zhou, S., Tang, W., Zheng, N., Hua, G.: Transfer Easy to Hard: Adversarial Contrastive Feature Learning for Unsupervised Person Re-Identification. Pattern Recognition 145, 109973 (2024)

28. Li, H., Hu, Q., Hu, Z.: Catalyst for Clustering-Based Unsupervised Object Re-Identification: Feature Calibration. In: Proceedings of the AAAI Conference on Artificial Intelligence, vol. 38, no. 4, pp. 3091–3099 (2024)

29. Liu, D., Fu, Y., Shi, W., Zhu, Z., Wang, D.: The Double Contrast for Unsupervised Person Re-Identification. In: Proceedings of the International Conference on Computer Vision and Deep Learning, pp. 1–8 (2024)